%% file: main.tex
\documentclass{article}

\usepackage[final]{corl_2026} 

\title{EventVLA: Event-Driven Visual Evidence Memory for Long-Horizon Vision-Language-Action Policies}

\usepackage{microtype}
\usepackage{graphicx}
\usepackage{subcaption}
\usepackage{booktabs} 
\usepackage{tabularx}

\usepackage{hyperref}

\usepackage{amsmath}
\usepackage{amssymb}
\usepackage{mathtools}
\usepackage{amsthm}
\usepackage{enumitem}

\usepackage[table]{xcolor}   
\usepackage{color}
\usepackage[most]{tcolorbox} 

\usepackage[capitalize,noabbrev]{cleveref}

\theoremstyle{plain}

\theoremstyle{definition}

\theoremstyle{remark}

\usepackage[textsize=tiny]{todonotes}
\definecolor{darkgreen}{RGB}{0,155,0}

\usepackage{graphicx}
\usepackage{CJK}
\usepackage{wrapfig}
\usepackage{booktabs}

\usepackage{makecell}
\usepackage{multirow}

\makeatletter
\def\blfootnote{\xdef\@thefnmark{}\@footnotetext}
\makeatother

\newcommand{\ours}{{EventVLA}}
\newcommand{\oursbench}{{RoboTwin-MeM}}

\usepackage{tabularx}

\author{
    \normalsize
    \textbf{Ganlin Yang$^{1,2*}$, Zhangzheng Tu$^{4,3*}$, Yuqiang Yang$^{2*}$, Sitong Mao$^{5}$, Junyi Dong$^{5}$} \\
    \textbf{Tianxing Chen$^{6}$, Jiaqi Peng$^{7,2}$, Jing Xiong$^{8,2}$, Jiafei Cao$^{2}$, Jifeng Dai$^{7}$, Wengang Zhou$^1$} \\
    \textbf{Yao Mu$^{3,2,6\dag}$, Tai Wang$^{2\dag}$} \\[2mm]
    \small 
    $^1$University of Science and Technology of China \quad $^2$Shanghai AI Laboratory \\ 
    \small
    $^3$Shanghai Jiao Tong University \quad $^4$Dalian University of Technology \quad $^5$Huawei Technologies Co., Ltd. \\ 
    \small
    $^6$The University of Hong Kong \quad $^7$Tsinghua University \quad $^8$Peking University \\[3mm]
    \normalsize
    Project Page: \href{https://ganlin-yang.github.io/EventVLA.github.io/}{EventVLA}
}

\begin{document}
\maketitle

\blfootnote{\noindent$^{*}$Equal contribution. $^{\dag}$Corresponding authors.}

\begin{abstract}
Memory remains a critical bottleneck for long-horizon robotic manipulation, as standard Vision-Language-Action (VLA) policies often fail when task-relevant cues become occluded or unobservable over time. While existing memory-augmented methods utilize historical context, they either suffer from severe information bottlenecks, incur high latency via decoupled dual systems, or rely on unselective buffers that accumulate massive visual redundancies. To address these limitations, we introduce \textbf{\ours}, an end-to-end framework founded on the concept of \textbf{sparse visual evidence memory} that comprises two core components: foundational \textit{visual anchors} to retain initial and short-term contexts, and a dynamic \textit{Keyframe Evidence Memory (KEM)} module. Specifically, KEM directly predicts future keyframe probabilities from the VLA's latent embeddings to autonomously capture and store sparse, task-critical visual events. This foresight-driven mechanism empowers the policy to dynamically evaluate the future causal utility of current observations, preserving transient visual evidence before it becomes unobservable. Furthermore, we propose \textbf{\oursbench}, a diagnostic benchmark specifically designed to evaluate non-Markovian manipulation tasks with interactive visual evidence. Extensive evaluations show that across 17 memory-requiring simulation tasks and 4 real-world bimanual tasks, \ours~achieves an average success rate improvement of +40\% over state-of-the-art memory-augmented VLAs. The code, models and datasets are available at \url{https://github.com/InternRobotics/EventVLA}.
\end{abstract}


\section{Introduction}

    Recent Vision-Language-Action (VLA) policies excel in generalizable and fine-grained manipulation~\cite{pi05, pi06, diffusionpolicy, 3ddp, zhao2023learning}, yet they predominantly operate under a strict Markovian assumption. This implicitly assumes all task-relevant information remains persistently visible. In reality, physical workspaces change dynamically, and agents must constantly retain intermediate states, such as the original location of a displaced item to guide subsequent actions. To address this non-Markovian challenge, memory-aware VLAs have emerged across three paradigms~\cite{dai2026robomme}. First, Dual-system Memory-VLAs~\cite{memer, chen2026rmbench, pimem} decouple cognition from control but suffer from high latency and severe error propagation. Second, Recurrent architectures~\cite{xiao2025ava, bulatov2022recurrent} compress history into hidden states, creating an information bottleneck that discards fine-grained visual details. Third, Memory Buffers~\cite{memoryvla, cronusvla} preserve visual fidelity but blindly accumulate redundant frames without a selective mechanism. Consequently, a critical question remains: exactly \textit{when} and \textit{what} visual evidence should a VLA preserve to maximize execution success without overwhelming computational limits?

\begin{figure*}[t]
    \centering
    \includegraphics[width=\textwidth]{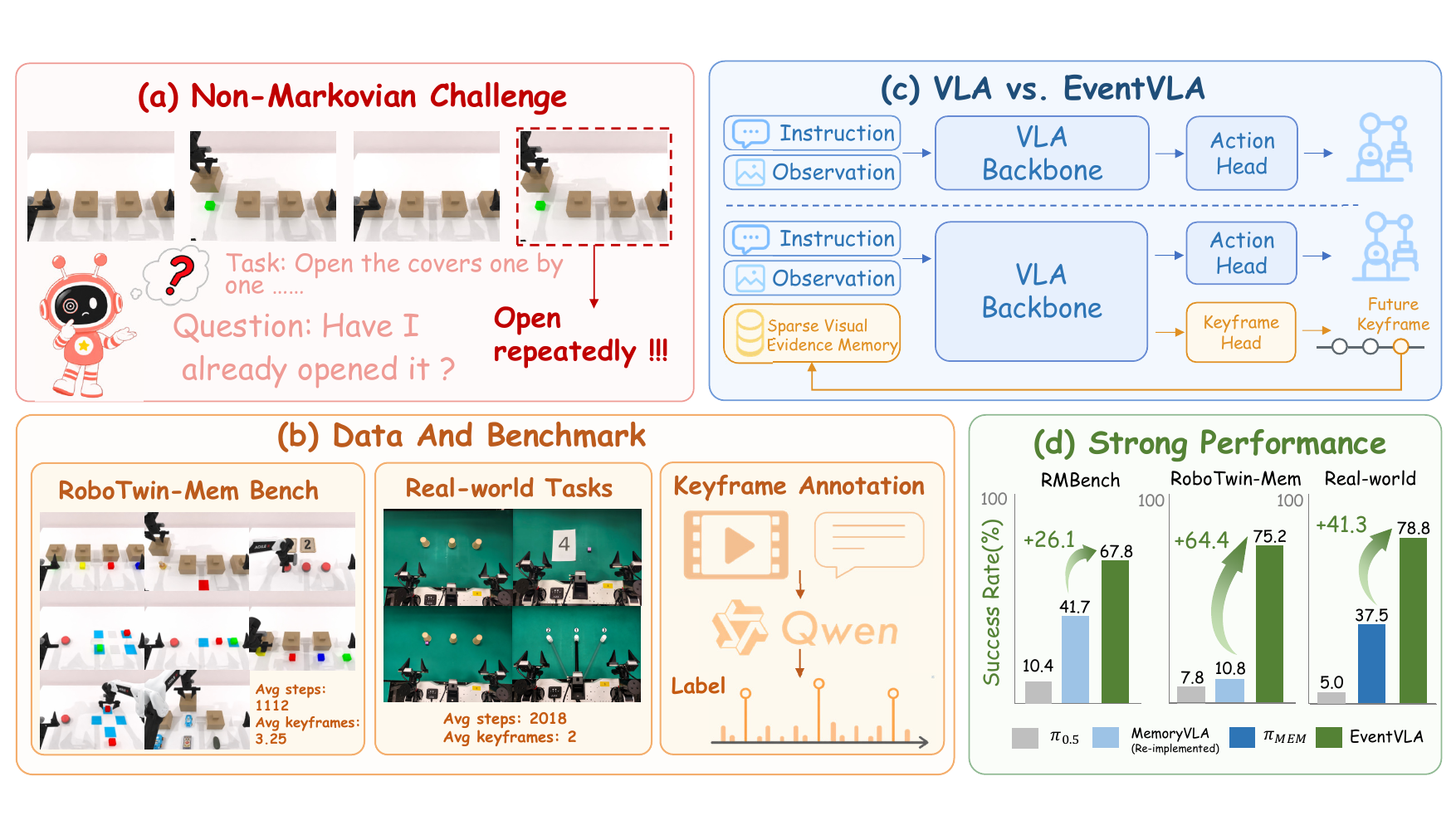}
    \caption{
    \textbf{Overview of EventVLA.} EventVLA tackles long-horizon, memory-requiring manipulation tasks by storing sparse, task-critical visual evidence. The figure illustrates the (a) non-Markovian challenge, (b) our proposed and evaluated benchmarks, (c) event-driven memory design, and (d) strong gains across simulation and real-world tasks.
    }
    \label{fig:overview}
\end{figure*}

To avoid the massive redundancy of standard memory buffers~\cite{memoryvla, wang2025lola}, we identify that a sparse set of historical keyframes provides sufficient context for many long-horizon tasks. We define these as foundational \textbf{visual anchors}: the initial frame (capturing the invariant global layout) and a short-term history window (providing local motion cues). While these heuristic anchors efficiently solve structurally simple memory-requiring tasks, they fundamentally fail in complex interactive scenarios where task-critical evidence emerges unexpectedly and subsequently disappears. For example, a robot may briefly observe an object's color when lifting an opaque cover, or need to track a designated target that later becomes occluded. Such transient visual evidence cannot be recovered from initial or recent frames; it manifests as an interactive \textit{sparse event} that must be actively captured and preserved.

Building upon this insight, we introduce \textbf{\ours}, an end-to-end framework rooted in sparse visual evidence memory. \ours~eliminates historical redundancy by seamlessly combining foundational visual anchors with a dynamic \textbf{Keyframe Evidence Memory (KEM)} module. Unlike rigid, rule-based heuristics, KEM establishes an autonomous, data-driven mechanism designed to actively capture transient, interaction-driven events. Specifically, by performing foresight-driven keyframe predictions over the upcoming execution horizon, KEM empowers the VLA policy to proactively schedule sparse memory writes for critical intermediate states. This predictive strategy ensures that transient visual evidence is captured long before it becomes explicitly required by the task, seamlessly bridging the temporal gap between its brief appearance and its eventual use in downstream execution. To learn this capability without prohibitive manual annotation, we develop an offline, Qwen3-VL-based~\cite{bai2025qwen3} automatic labeling pipeline that extracts precise keyframe supervision from demonstrations.


Beyond algorithm design, evaluating memory-augmented policies requires benchmarks that accurately capture the non-Markovian dynamics in real-world manipulation, where task-critical evidence often manifests only transiently during intermediate interactions. Because existing benchmarks like RMBench~\cite{chen2026rmbench} can largely be solved by basic visual anchors alone, we introduce \textbf{\oursbench}, a diagnostic simulation benchmark explicitly featuring such genuinely non-Markovian scenarios. It comprises 8 challenging tasks, where the required intermediate keyframes systematically scale from 1 to 5. Extensive evaluations demonstrate \ours's superiority across diverse domains. It sets a new state-of-the-art on conventional memory-oriented tasks (67.8\% on RMBench) and achieves a 75.2\% average success rate on the newly transient-memory-required \oursbench, vastly outperforming existing memory-based VLAs. Furthermore, in demanding real-world bimanual tasks, \ours~significantly surpasses both reactive ($\pi_{0.5}$~\cite{pi05}) and memory-augmented ($\pi_{MEM}$~\cite{pimem}) baselines with up to 90\% success rates, confirming its robust non-Markovian situational awareness.




\section{Related Work}

\subsection{Memory-Augmented Policies for Long-Horizon Manipulation}

Recent Vision-Language-Action (VLA) foundation models~\cite{pi05,pi06,rdt2,xvla,zhao2023learning,diffusionpolicy,3ddp,g3flow,dexvla,hifvla,discretediffusionvla,yang2025vlaser,shen2025expertise,diffusionvla,tinyvla,yang2026visual} achieve remarkable generalizability but are fundamentally memoryless. Operating under a strict Markovian assumption, they struggle with non-Markovian tasks where critical visual information is transient or occluded. To address this, memory-augmented VLAs have emerged across three paradigms. First, \textit{dual-system Memory-VLAs}~\cite{memer, chen2026rmbench, pimem, zheng2025tracevla, tan2026action} use a high-level VLM for planning but suffer from error propagation and high inference latency. Second, \textit{recurrent memory architectures}~\cite{xiao2025ava, bulatov2022recurrent, vqmemory, longcontextdp, cyclemanip} compress histories into hidden states, creating an information bottleneck that discards fine-grained visual details. Third, \textit{Memory Buffers}~\cite{contextvla, memoryvla, cronusvla, echovla, vpwem, memoact, wang2025lola} retain historical frames to bypass compression; however, existing methods blindly accumulate redundant frames, drowning out sparse key evidence and incurring heavy overhead. \ours~optimizes this paradigm by preserving only sparse visual evidence. By combining static \textit{visual anchors} with a dynamic \textit{Keyframe Evidence Memory (KEM)}, \ours~selectively captures transient states, balancing robust task execution with real-time computational efficiency.
Different from prior retrieval-based keyframe memory such as MemER~\cite{memer}, our contribution lies in foresight-based, end-to-end future keyframe prediction rather than keyframe memory itself.

\subsection{Memory-Oriented Manipulation Benchmarks}

Standard simulation suites~\cite{robotwin2.0, robocasa, maniskill3, behavior1k, li2024evaluating} emphasize long-horizon execution rather than explicit memory reasoning, as task-relevant information typically remains persistently visible. While recent memory-centric benchmarks~\cite{sam2act, MIKASA, libero, robocerebra, lei2026robomemarena} attempt to address this, they are often limited in scale, tailored exclusively for reinforcement learning, or still feature observable states. 
The closest suites to ours, RMBench~\cite{chen2026rmbench} and RoboMME~\cite{dai2026robomme}, systematically stratify memory demands. RMBench can often be addressed using static visual anchors, whereas RoboMME evaluates broader memory capabilities.
To bridge this methodological gap, we introduce \textbf{\oursbench}. Distinct from existing benchmarks, \oursbench~isolates genuinely non-Markovian manipulation tasks where critical visual evidence transiently emerges during interaction and subsequently disappears, providing a rigorous diagnostic platform to evaluate a VLA policy's capacity for intermediate state retention.

\section{EventVLA Framework}
\label{sec:eventvla}

\subsection{Problem Formulation and Foundational Visual Anchors}

We formalize long-horizon robotic manipulation as a non-Markovian decision process. Standard reactive VLA policies map the current observation $o_t$ and language instruction $l$ directly to an action, i.e., $a_t = \pi(o_t, l)$, which fundamentally fails when critical information becomes occluded or unobservable over time. To address this, EventVLA incorporates an explicit, external sparse visual evidence memory buffer $M_{t}$ to condition action generation along with the immediate observation:
\begin{equation}
a_t = \pi(o_t, M_{t-1}, l)
\end{equation}
where $M_{t-1}$ selectively stores key historical frames to preserve essential visual evidence while minimizing informational and computational redundancy.

\begin{wrapfigure}{r}{0.6\textwidth}
    \centering
    \includegraphics[width=1.05\linewidth]
    {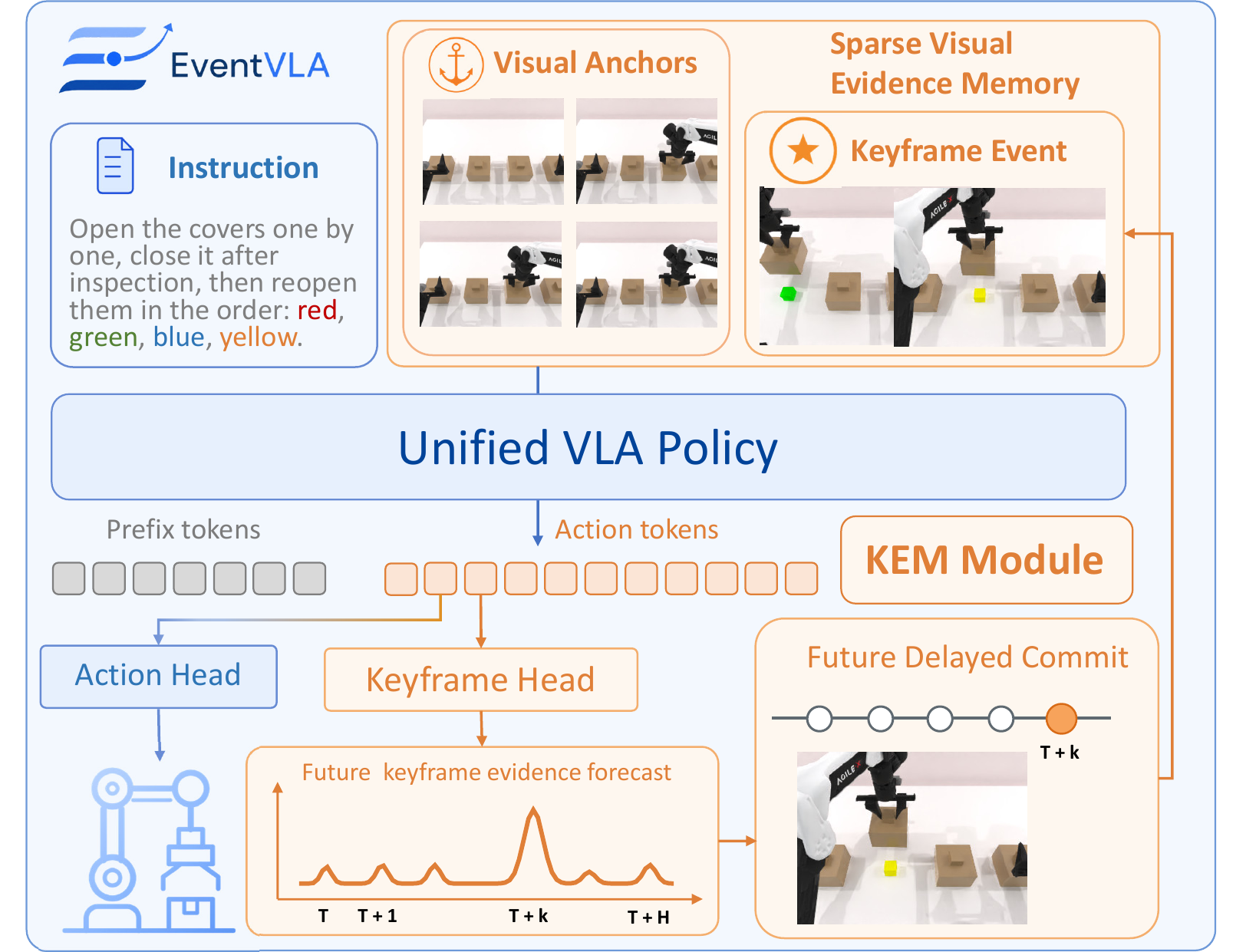}
    {
    \caption{ 
    \textbf{EventVLA framework.} EventVLA maintains a sparse visual evidence memory composed of foundational visual anchors and interaction-driven event keyframes, and uses the KEM module to proactively commit task-critical future key observations into memory. 
    }
    \label{fig:eventvla}
    }
\end{wrapfigure}

The memory buffer is structured as $M_t = A_t \cup E_t$, seamlessly uniting foundational \textit{visual anchors} $A_t$ and interaction-driven event keyframes $E_t$. The visual anchors $A_t$ represent a deterministic, rule-based baseline designed to capture the permanent scene layout and immediate temporal context. Specifically, the visual anchors at timestep $t$ consist of the initial workspace configuration $o_0$ and a short-term history sliding window of size $K$:
\begin{equation}
A_t = {o_0} \cup {o_{t-K}, \dots, o_{t-1}}
\end{equation}
Here, the initial frame $o_0$ serves as a permanent spatial anchor, allowing the VLA model to preserve an invariant memory of the original scene arrangement before any displacements occur. Meanwhile, the short-term history $o_{t-i}$ supplies the model with critical motion and task progression cues, enabling smooth and continuous action generation. However, since these rigid anchors cannot capture unpredictable, transient evidence arising midway through complex interactions, they are dynamically augmented by $E_t$ produced by the Keyframe Evidence Memory (KEM) module, as detailed in Section~\ref{sec_kem} and the overall framework is shown in Fig.~\ref{fig:eventvla}.

\subsection{Keyframe Evidence Memory (KEM) Module}
\label{sec_kem}

To actively capture transient, interaction-driven events that foundational visual anchors inherently miss, such as the brief exposure of an occluded object, we introduce the Keyframe Evidence Memory (KEM) module. To implement this mechanism efficiently, KEM is designed as a lightweight, parallel prediction head operating directly alongside the primary action heads. Rather than utilizing isolated features, the keyframe prediction head ingests the exact hidden states $h_t \in \mathbb{R}^{H \times d}$ extracted from the final layer of the VLA's autoregressive transformer, for action horizon $H$. Because $h_t$ naturally encapsulates the joint embedding of visual observations and action-conditioned query tokens, the keyframe head inherits a proactive awareness of the model's future execution plan. Specifically, the keyframe head projects these shared hidden states $h_t$ to a vector of keyframe probabilities $\hat{\mathbf{p}}_t$ spanning the future chunk horizon $H$:
\begin{equation}
\hat{\mathbf{p}}_t = \sigma(\text{KEM}_{\text{mlp}}(h_t)) = [\hat{p}_t^1, \hat{p}_t^2, \dots, \hat{p}_t^H]^T \in [0, 1]^H
\end{equation}
where $\sigma(\cdot)$ denotes the element-wise sigmoid function, and each scalar $\hat{p}_t^i \in [0, 1]$ explicitly represents the predicted probability of the $i$-th future execution step being a task-critical keyframe. 
For the controlled ablation, we use $H_k \leq H$ to denote the evaluated keyframe foresight horizon while keeping the action horizon $H$ fixed; the default setting is $H_k=H=50$.
The rationale for this chunk-wise prediction is straightforward: a purely step-wise classifier would completely miss task-critical events that transiently manifest and vanish midway through the execution window (e.g., at step $t+i$ where $0 < i < H$). This limitation motivates KEM to adopt a foresight-driven, chunk-wise paradigm $\hat{\mathbf{p}_t}$, empowering the VLA policy to proactively map out a ``memory schedule" across the entire upcoming execution horizon.

Driven by this predictive vector $\hat{\mathbf{p}}_t$, \ours~triggers a sparse memory write \textit{event} whenever a predicted probability crosses a threshold ($\hat{p}_t^i \geq \tau_{\text{commit}}$), dynamically committing the raw image at $t+i$ to the event buffer $E_t$. To satisfy real-time constraints, $E_t$ is bounded by a maximum capacity $N_{\text{max}}$, managed via a First-In-First-Out (FIFO) eviction policy. At any execution step $t$, these dynamically accumulated event keyframes $E_{t-1}$ are seamlessly combined with the foundational visual anchors $A_t$ and the immediate observation $o_t$ into a single, temporally ordered sequence:
\begin{equation}
I_{\text{input}} = concatenate([A_{t}, E_{t-1}, o_t])
\end{equation}
Feeding this unified sequence directly into the VLM's vision encoder allows the self-attention layers to dynamically extract complex temporal correlations across sparse historical frames, natively endowing the model with robust situational awareness for long-horizon manipulation.

\subsection{End-to-End Training and Inference Details}

To train the KEM module without prohibitive manual annotation costs, we employ an offline Qwen3-VL~\cite{bai2025qwen3} automated pipeline to extract ground-truth timestamps of task-critical events. To mitigate the inherent temporal ambiguity of physical interactions, we supervise chunk-wise keyframe predictions using temporally smoothed soft labels via a sequence-averaged BCE objective ($L_{\text{kem}}$). The framework is optimized end-to-end alongside the standard action generation loss ($L_{\text{action}}$):
\begin{equation}
L = L_{\text{action}} + \lambda L_{\text{kem}}
\end{equation}
To bridge the train-test distribution shift while maintaining early training stability, we apply a scheduled teacher-to-student curriculum that gradually transitions memory construction from ground-truth to autonomous predictions.
During online inference, continuous keyframe probabilities naturally cluster around unfolding semantic events. To enforce strict memory sparsity and prevent redundant buffer flooding, we distill these dense predictions into discrete write events using a 1D Non-Maximum Suppression (NMS) and temporal cooldown pipeline. Comprehensive mathematical formulations regarding the soft labels, curriculum, NMS algorithm, and the automated labeling pipeline are deferred to Appendix~\ref{appden:eventvla_details}. Additionally, complete network structures and training configurations are detailed in Appendix~\ref{appen:hyperparas}.

\section{\oursbench~Benchmark}
\label{sec:memdojobench}

\begin{figure*}[t]
    \centering
    \includegraphics[width=0.95\linewidth]{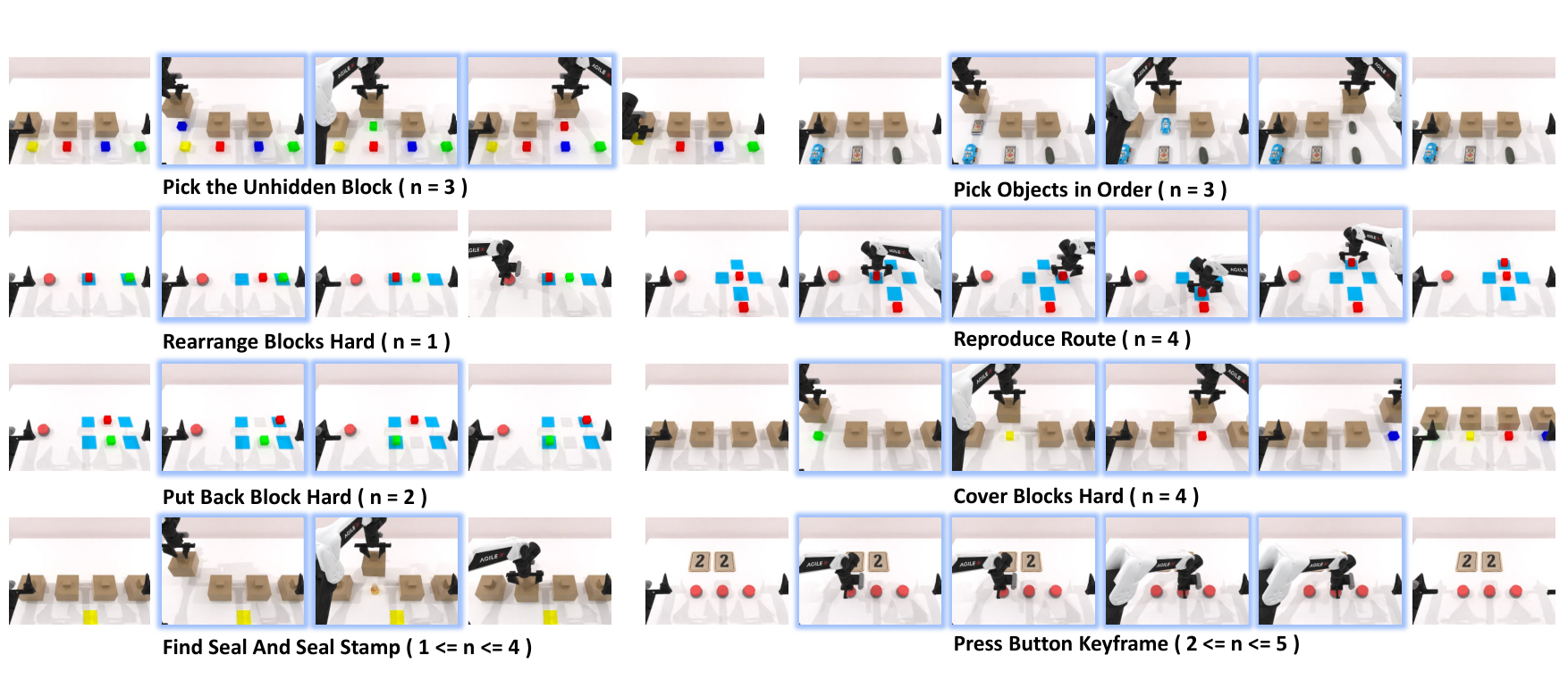}
    {
    \vspace{-0.5em}
    \caption{ 
    \textbf{Overview of the 8 evaluation tasks in the \oursbench~benchmark.} To rigorously evaluate the capacity for intermediate visual evidence retention, each task is explicitly parameterized by $n$ (ranges from 1 to 5), denoting the exact number of transient, interaction-driven keyframes that must be memorized to succeed. These task-critical intermediate events are highlighted with blue borders. 
    }
    \label{fig:memdojo}
    }
\end{figure*}

To systematically evaluate the capability of VLA policies to capture and retain transient visual evidence, we introduce \oursbench, a diagnostic simulation benchmark. Developed within the RoboTwin~2.0~\cite{robotwin2.0} simulation platform and built on top of the SAPIEN~\cite{sapien} physics engine, \oursbench~supports both automated data synthesis and integrated policy evaluation within a unified pipeline. This infrastructure ensures scalable data generation alongside consistent, reproducible benchmarking for robotic manipulation. Furthermore, we provide fine-grained language annotations that align strictly with each action-observation pair. These annotations assign explicit linguistic descriptions to low-level interactions and state transitions, offering structured and dense supervision signals that are highly beneficial for training downstream memory modules.

The core distinction between \oursbench~and existing memory-centric suites is its explicit isolation and quantification of intermediate memory demands. Rather than introducing entirely different task semantics, \oursbench~provides a controlled extension of RMBench-style tasks in which task-critical visual evidence emerges during execution and later becomes unobservable.
To rigorously diagnose this capability, we explicitly parameterize task complexity using $n$: the exact number of intermediate event keyframes that must be dynamically preserved. As illustrated in Fig.~\ref{fig:memdojo}, \oursbench~comprises 8 genuinely non-Markovian tasks featuring extremely long execution horizons, averaging between 430 and 1544 steps per episode. Across the benchmark, the required intermediate keyframe count $n$ systematically ranges from 1 to 5, establishing a tiered difficulty hierarchy for non-Markovian control. The detailed task statistics and language instructions can be found in Table~\ref{memdojo}.

Crucially, this $n$-parameterized design allows \oursbench~to evaluate a diverse spectrum of memory capabilities beyond trivial history concatenation. First, tasks like \textit{Pick the Unhidden Block} ($n=3$) and \textit{Cover Blocks Hard} ($n=4$) demand \textit{transient memory}; essential visual evidence is briefly exposed when a cover is lifted and completely disappears once closed, requiring the policy to instantly anchor this fleeting information. Second, the benchmark evaluates sequence tracking and counting logic via tasks like \textit{Press Button Keyframe} ($n \in [2, 5]$), where each button press represents an execution-critical event that must be sequentially registered to dictate task success. Finally, the \textit{Reproduce Route} task ($n=4$) tests the model's \textit{in-context learning} capacity, requiring the agent to observe a demonstration, extract randomized spatial keypoints, and leverage these cues in-context to duplicate the route. This coverage of transient recognition, event counting, and in-context imitation makes \oursbench~a rigorous benchmark for evaluating memory-augmented robotic policies.

\section{Experiments}

\subsection{Performance on Simulation Benchmarks}

\input{table/rmbench_abbr}

To thoroughly assess the efficacy of \ours, we benchmark our framework against a comprehensive suite of state-of-the-art baselines, categorized into two major paradigms. For standard, reactive (non-memory-based) VLA policies, we select DP~\cite{diffusionpolicy}, ACT~\cite{zhao2023learning}, $\pi_{0.5}$~\cite{pi05}, X-VLA~\cite{xvla}, and QwenOFT~\cite{starvla}. For memory-augmented methods, we evaluate dual-system architectures, including MemER~\cite{memer} and Mem-0~\cite{chen2026rmbench}, as well as the end-to-end MemoryVLA~\cite{memoryvla} framework, where we reproduce its variants based on both the official OpenVLA-OFT~\cite{oft} and QwenOFT~\cite{starvla} implementations. Our proposed \ours~is also constructed upon the identical open-source QwenOFT backbone as its foundational base model.

\noindent\textbf{Evaluation on RMBench and the Efficacy of Visual Anchors:}
First, we evaluate our method on RMBench~\cite{chen2026rmbench}, as shown in Table~\ref{benchmark:rmbench_abbr}. Because tasks in this suite primarily rely on persistent spatial layouts and fixed motion style rather than hidden intermediate states, we deploy a streamlined version of \ours~utilizing solely foundational visual anchors. Experimental results demonstrate that this configuration achieves an average success rate of 67.8\%, securing state-of-the-art performance and proving that rule-based anchors provide sufficient context for simple memory-required long-horizon manipulation. To validate the structural necessity of these components, we conduct two ablation studies. Removing the initial frame (\ours~w/o initial) or discarding the short-term history (\ours~w/o short-term) causes the overall success rate to plummet to 33.7\% and 23.8\%, respectively. This confirms that both the initial global spatial reference and the short-term motion cues are indispensable for effective visual anchoring.

\input{table/rmbench_new}

\textbf{Evaluation on \oursbench~and the Necessity of KEM:}
While foundational visual anchors excel on RMBench, their limitations become starkly apparent when evaluated on \oursbench, our diagnostic suite explicitly designed to test intermediate state memory. As detailed in Table~\ref{benchmark:memdojo}, relying solely on rule-driven visual anchors (VA only) yields a mere 18.0\% average success rate. This sharp drop indicates that fixed historical windows are fundamentally inadequate for tasks requiring VLA policies to retain transient visual evidence generated mid-execution. To overcome this non-Markovian bottleneck, the full EventVLA framework augments these visual anchors with the dynamic Keyframe Evidence Memory (KEM) module (VA+KEM). Experimental results reveal a qualitative leap: the complete EventVLA achieves a 75.2\% success rate, outperforming all baseline models by a substantial margin. This striking performance delta (from 18.0\% to 75.2\%) compellingly demonstrates that KEM's dynamic event capture and foresight-driven writing mechanisms are indispensable for solving complex, long-horizon tasks that hinge on transient intermediate memory.

\input{table/robotwin50}
\noindent\textbf{Evaluation on Standard Markovian Benchmarks:}
To verify that \ours~preserves fundamental reactive control, we evaluate it on standard Markovian tasks in RoboTwin-2.0~\cite{robotwin2.0} (Table~\ref{benchmark:robotwin2.0}). Rather than degrading performance, our memory mechanism slightly improves success rates over the memoryless QwenOFT baseline (80.0\% to 83.8\% on Easy; 78.0\% to 81.6\% on Hard), seamlessly complementing standard closed-loop execution.

\subsection{Ablation Analysis of \ours~}
\label{sec:eventvla_abla}
\input{table/rebuttal_memory}

To systematically validate the structural design of the Keyframe Evidence Memory (KEM) module, we conduct ablation studies on the challenging \oursbench~suite (Table~\ref{benchmark:memdojo}, bottom in gray). 

\textbf{Core Mechanisms:} We observe that replacing explicit raw image concatenation with an implicit latent memory bank drastically drops the success rate from 75.2\% to 24.9\%, creating a severe information bottleneck. Similarly, substituting temporally smoothed soft labels with rigid binary targets destabilizes the predictive head, reducing performance to 48.8\%. 

\textbf{Buffer and Horizon Management:} Removing the NMS post-processing or restricting the buffer capacity ($N_{\text{max}}=2$) leads to redundant frame flooding and premature FIFO eviction of critical early evidence, degrading success rates to 53.4\% and 32.0\%, respectively. 
To isolate keyframe foresight from the effect of action-chunk length, we fix the action horizon at $H=50$ and vary only the keyframe foresight horizon $H_k$. As detailed in Appendix~\ref{appendix:abla}, extending $H_k$ from 1 to 50 progressively improves performance from 28.2\% to 75.2\% on \oursbench, while the reactive selector ($H_k=0$) achieves only 28.9\%.

Table~\ref{tab:rebuttal_memory} further compares memory construction strategies under the same backbone and memory budget. On \oursbench, \ours~substantially outperforms Random, Periodic, Visual-change, and Reactive selectors, and approaches the oracle-keyframe upper bound. The component results further show that KEM provides the major gains on tasks requiring transient intermediate evidence. On RoboMME, VA+KEM improves over VA only from 16.4\% to 32.1\%, confirming that the benefit of KEM is not specific to \oursbench.
Comprehensive in-depth analyses of these ablation modes, along with real-time inference speed profiling, are deferred to Appendix~\ref{appendix:abla}.

\begin{figure*}[t]
    \centering
    \includegraphics[width=0.95\linewidth]{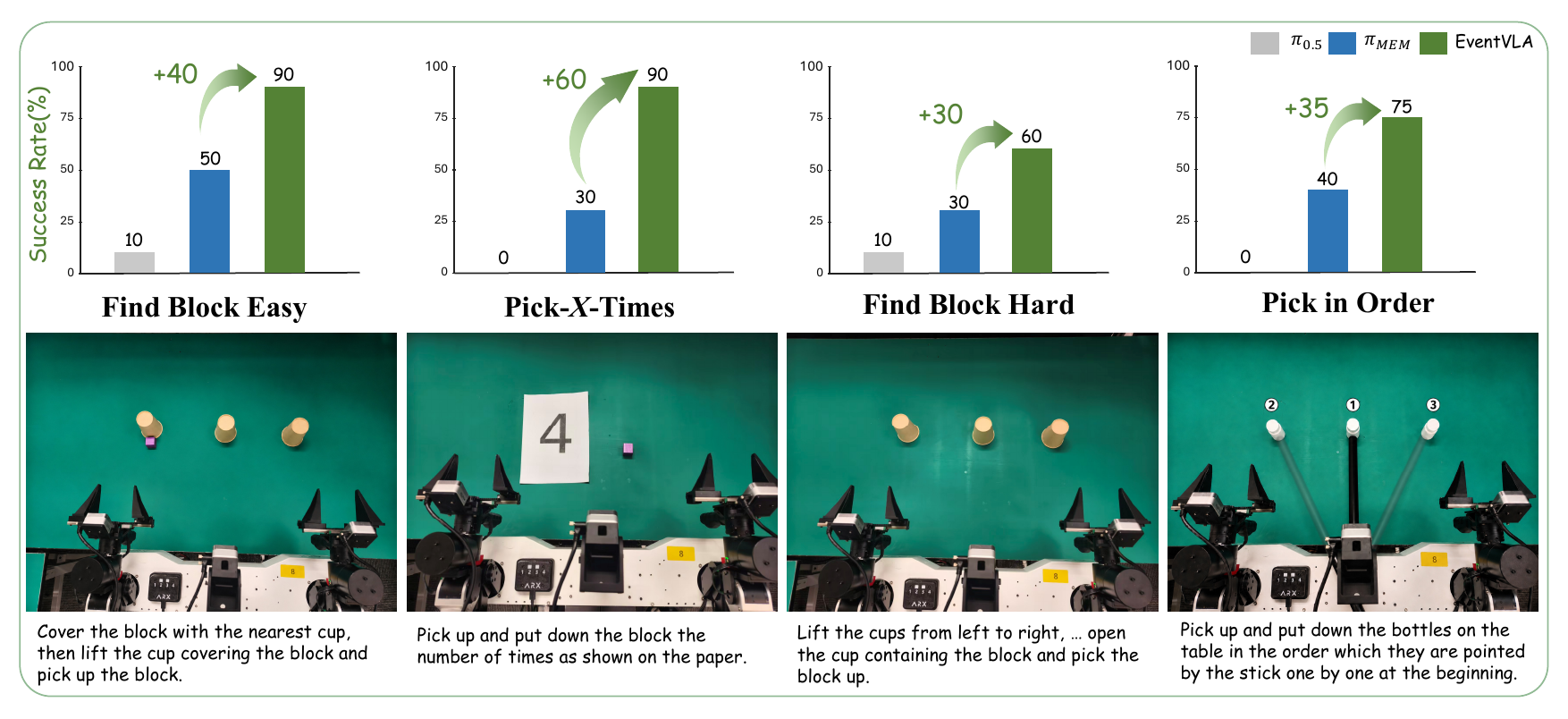}
    {
    \caption{ 
    Real-world experimental setups and results on the ARX ACONE bimanual robot. We evaluate four memory-intensive manipulation tasks: \textit{Find Block Easy}, \textit{Find Block Hard}, \textit{Pick-X-Times}, and \textit{Pick in Order}.
    }
    \label{fig:real_robot}
    }
\end{figure*}

\subsection{Real-World Robot Evaluation}
To evaluate \ours~in physical environments, we deploy our framework on the ARX ACONE bimanual robot across four non-Markovian manipulation tasks, each tested over 20 independent trials. These tasks explicitly evaluate diverse cognitive memory capabilities under real-world settings: 1) \textbf{\textit{Find Block Easy}} and \textbf{\textit{Find Block Hard}} require the model to remember the spatial location of a hidden block after only transient visual exposure. 2) \textbf{\textit{Pick-X-Times}} tests counting logic, requiring the robot to read a randomized number and manipulate a block accordingly. 3) \textbf{\textit{Pick in Order}} evaluates in-context memory by asking the robot to reproduce a randomized sequence initially pointed out by a stick. We benchmark EventVLA against the non-memory $\pi_{0.5}$ policy~\cite{pi05}, the reproduced memory-augmented $\pi_{MEM}$ baseline~\cite{pimem}, and the keyframe-based MemER baseline~\cite{memer}.

As illustrated in Fig.~\ref{fig:real_robot}, the purely reactive $\pi_{0.5}$ policy almost entirely fails across all tasks (achieving only 0\% to 10\% success rates) as it fundamentally lacks the historical context required to infer occluded states.
While the memory-augmented $\pi_{MEM}$ baseline demonstrates partial improvements, its dual-system execution must repeatedly detect subtask completion and transition across decomposed cycles, such that missed or additional transitions can accumulate into task failure. The keyframe-based MemER baseline achieves an average success rate of 32.3\% (Table~\ref{tab:inference_cost}). In contrast, \ours~achieves success rates of 90\%, 90\%, 60\%, and 75\% on \textit{Find Block Easy}, \textit{Pick-X-Times}, \textit{Find Block Hard}, and \textit{Pick in Order}, respectively, corresponding to an average success rate of 78.8\%.
This robust physical performance validates that the KEM module can effectively extract and retain critical transient visual cues, empowering the VLA model with long-horizon memory awareness in the real world.

\section{Limitations}
While \ours~effectively captures transient visual evidence, its bounded event buffer limits scalability in exceptionally long-horizon tasks (e.g., $>10$ minutes) with high event densities. Such scenarios risk buffer saturation and premature eviction of early historical cues. Future work will explore hierarchical memory or compressed representations to manage massive event sequences.
In addition, cross-task transfer of learned memory behavior to unseen tasks, objects, or scene distractors remains unexplored and is left for future work.

\section{Conclusion}
We introduced \ours, an end-to-end framework tackling non-Markovian long-horizon manipulation via sparse visual evidence memory. By uniting rule-based visual anchors with a foresight-driven Keyframe Evidence Memory (KEM) module, \ours~proactively captures task-critical transient events, completely avoiding the redundancy of dense memory. Furthermore, we proposed \oursbench, a diagnostic benchmark for evaluating intermediate memory capabilities. Extensive evaluations across 17 simulation and 4 real-world tasks demonstrate that \ours~significantly outperforms state-of-the-art memory-augmented VLAs, ensuring robust memory-requiring long-horizon physical execution.

\acknowledgments{This work is supported by Shanghai Artificial Intelligence Laboratory.
This work was supported in part by the National Natural Science Foundation of China under Grant No. 62606316 and the Shanghai Magnolia Talent Program Pujiang Project under Grant No. 25PJA076.}


\bibliography{example}  

\newpage
\appendix

\section*{Appendix}

\section{Implementation Details of EventVLA}
\label{appden:eventvla_details}

\subsection{Training Formulations and Curriculum Strategy}

To obtain ground-truth (GT) keyframe supervisions, we leverage an offline automated labeling pipeline powered by Qwen3-VL~\cite{bai2025qwen3}. By parsing raw demonstration videos alongside task descriptions, the VLM extracts the exact timestamps of task-critical intermediate events, denoted as $t^*$. However, in physical robot execution, keyframe semantics inherently exhibit temporal ambiguity—frames immediately preceding or succeeding $t^*$ are often equally valid for capturing the visual evidence. To prevent noisy gradients caused by rigid binary supervision, we smooth the annotations into a soft target vector $\mathbf{y}_t \in [0, 1]^H$ utilizing a raised cosine kernel. Specifically, for a future step $i$ within a dilation radius $R$ of a GT event $t^*$, the soft target is defined as $y_t^i = 0.5 (1 + \cos(\pi \frac{|t+i - t^*|}{R}))$.

To supervise the chunk-wise keyframe predictions against these temporally smoothed annotations, we formulate the Keyframe Evidence Memory loss $L_{\text{kem}}$ as a sequence-averaged Binary Cross-Entropy (BCE) objective. This explicitly aligns each predicted scalar probability $\hat{p}_t^i \in [0, 1]$ with its corresponding soft target $y_t^i \in [0, 1]$ across the entire future action horizon $H$:
\begin{equation}
L_{\text{kem}} = -\frac{1}{H} \sum_{i=1}^{H} \left[ y_t^i \log(\hat{p}_t^i) + (1 - y_t^i) \log(1 - \hat{p}_t^i) \right]
\end{equation}
The entire framework is then optimized end-to-end via a joint objective that couples memory awareness with precise motor control:
\begin{equation}
L = L_{\text{action}} + \lambda L_{\text{kem}}
\end{equation}
where $L_{\text{action}}$ denotes the standard continuous action generation loss (e.g., regression or flow-matching), and $\lambda$ serves as a balancing coefficient to appropriately scale the memory supervision.

During training, constructing the event buffer $E_t$ dynamically from the model's own predictions in the early stages causes severe training instability, whereas relying exclusively on GT keyframes introduces a critical train-test distribution shift since GT keyframes are unavailable at test time. To bridge this gap, we implement a scheduled teacher-to-student curriculum. We introduce an annealing parameter $\alpha$ that linearly decays from $1$ to $0$ over the training duration. At each step, the framework decides whether to commit an observation to $E_t$ using the GT keyframes with probability $\alpha$ (teacher-forcing), or relying on its own thresholded predictions ($\hat{p}_t^i \geq \tau_{\text{commit}}$) with probability $1-\alpha$. This gradual transition ensures stable initial convergence while forcing the VLA policy to eventually adapt to its own autonomous memory updating cadence.

\subsection{Online Inference and Post-Processing}
During online inference, the chunk-wise prediction $\hat{\mathbf{p}}_t$ naturally yields clustered, temporally continuous high-probability scores around an unfolding semantic event. To prevent redundant frames of the same visual event from flooding the bounded buffer $E_t$, we compress the $H$-dimensional probability vector into discrete, sparse write events via a rigorous post-processing extraction pipeline.

First, we identify a set of local probability peaks $\mathcal{K}_t$ by applying the confidence threshold $\tau_{\text{commit}}$ coupled with a 1D Non-Maximum Suppression (NMS) algorithm. Specifically, a future step index $i$ is selected as a candidate peak if its probability exceeds the threshold and represents the local maximum within a sliding temporal window of radius $w$:
\begin{equation}
\mathcal{K}_t = \left\{ i \in \{1, \dots, H\} \;\middle|\; \hat{p}_t^i \geq \tau_{\text{commit}} \land \hat{p}_t^i = \max_{j \in \mathcal{N}_w(i)} \hat{p}_t^j \right\}
\end{equation}
where $\mathcal{N}_w(i) = [\max(1, i-w), \min(H, i+w)]$ denotes the NMS neighborhood.

While NMS effectively isolates local peaks, rapid consecutive events might still trigger excessive memory writes. To strictly enforce operational sparsity, a temporal cooldown period $C$ is evaluated sequentially over the candidates. A candidate peak $i \in \mathcal{K}_t$ \textit{is officially validated and committed to $E_t$} if and only if $(t+i) - t_{\text{last}} > C$, where $t_{\text{last}}$ denotes the absolute physical timestamp of the most recently committed keyframe. Through this cascading extraction mechanism, the framework mathematically distills the dense predictive landscape into an optimal, highly sparse subset. This guarantees that memory allocation remains strictly tied to novel interactive evidence, maximizing information retention while seamlessly adhering to real-time execution constraints and bounded memory buffer size $N_{\text{max}}$.

\subsection{Automated Keyframe Annotation Pipeline}

To circumvent the prohibitive costs associated with dense manual frame annotation for long-horizon tasks, we develop an automated, highly scalable keyframe labeling pipeline powered by Large Vision-Language Models (VLMs). Specifically, we deploy the state-of-the-art \textbf{Qwen3-VL-235B-A22B-Instruct-FP8}~\cite{bai2025qwen3} model on a local server equipped with 8 NVIDIA A800 GPUs using the vLLM~\cite{vllm} framework.

\textbf{Data Pre-processing and In-Context Learning.} 
Rather than feeding an unmanageably long continuous video stream directly into the VLM, we uniformly sample the temporal horizon into a discrete set of frames (e.g., 128 frames per episode). To ensure robust spatial awareness, particularly in scenarios involving severe occlusions, we extract and concatenate multi-view observations (e.g., global head camera and wrist camera) for each sampled timestep. Crucially, to align the VLM's outputs with our specific definition of transient visual evidence, we employ an In-Context Learning (ICL) strategy. The prompt includes a few-shot demonstration from identical or similar tasks, containing the sampled frames alongside their ground-truth keyframe steps, establishing a rigorous template for temporal alignment and JSON-formatted outputs.

For each task, we manually annotate 2--3 training demonstrations as few-shot examples and use Qwen3-VL to annotate the remaining training trajectories. These few-shot examples are drawn only from the training split and never overlap with evaluation trajectories. Training and evaluation episodes use independently randomized object layouts and language instructions. At test time, neither keyframe annotations nor the required keyframe count is provided to the policy; KEM predicts and stores keyframes autonomously.

The exact system prompt utilized for the automated pipeline is presented below:

\begin{tcolorbox}[
    colframe=cyan!40!black,
    title=\textbf{System Prompt for VLM-based Keyframe Annotation},
    breakable,
    enhanced,
    left=2mm, right=2mm, top=2mm, bottom=2mm,
    fontupper=\footnotesize
]
\scriptsize
\textbf{System Prompt:} \\
You are an expert robot-video keyframe annotator.

\textbf{CRITICAL REQUIREMENT:} \\
- The annotation MUST strictly follow \texttt{task\_instruction}. \\
- Treat \texttt{task\_instruction} as the primary objective definition; if visuals are ambiguous, prioritize consistency with it.

\textbf{Episode metadata:} \\
- \texttt{episode\_id}: $<$episode\_id$>$ \\
- \texttt{task\_instruction}: $<$task\_instruction$>$ \\
- \texttt{total\_frames}: $<$total\_frames$>$ \\
- \texttt{required\_keyframe\_count}: $<$num\_keyframes$>$ \\
- \texttt{provided\_views}: $<$selected\_views$>$

\textbf{What to annotate:} \\
1. Find key state transitions for this task (e.g., stable grasp acquired, object placed, cycle transition). \\
2. Keep keyframes representative and temporally ordered across the full task progress. \\
3. For repeated pick/place cycles, pick the most stable and recognizable moments per cycle.

\textbf{Output format constraints:} \\
1. Return JSON only. No markdown, no explanations. \\
2. Format: \{ "keyframe\_steps": [int, int, ...] \} \\
3. \texttt{keyframe\_steps} must: \\
\hspace*{4mm} - have length exactly \texttt{<num\_keyframes>} \\
\hspace*{4mm} - be strictly increasing \\
\hspace*{4mm} - be in [0, \texttt{<total\_frames - 1>}] \\
\hspace*{4mm} - contain no duplicates
\end{tcolorbox}

\textbf{Annotation Reliability and Error Analysis.} 
To rigorously validate the reliability of this automated pipeline, we conducted a comprehensive cross-validation study. In the simulation environments, we compared the keyframes automatically annotated by the Qwen3-VL-235B model against the precise algorithmic ground-truth (GT) states acquired directly from the RoboTwin 2.0~\cite{robotwin2.0} physics engine. The results demonstrate that the VLM's predictions exhibit an average absolute temporal error of \textbf{less than 10 timesteps}. Furthermore, when deployed on the four complex real-world bimanual tasks, the prediction error remained \textbf{within 50 timesteps} compared to human-annotated ground truth. Given that our evaluation episodes feature extremely long horizons (often exceeding 1500 to 2000 steps), this negligible temporal variance, which is naturally accommodated by our temporally smoothed soft labels, strongly confirms that our VLM-powered automated annotation pipeline is highly reliable, precise, and ready for scalable deployment. 
We further evaluate the effect of annotation supervision on downstream policy performance. As reported in Table~\ref{tab:rebuttal_foresight}, Qwen3-VL supervision achieves 75.2\% on \oursbench~and 78.8\% in the real world, closely matching simulator-GT supervision (76.3\%) and human supervision (77.7\%), respectively.

\section{Experimental Setups and Benchmarks}

\subsection{\oursbench~Benchmark Details}

\input{table/memdojo_task}

As introduced in Sec.~\ref{sec:memdojobench} of the main text, \oursbench~is a diagnostic simulation suite specifically engineered to isolate and evaluate genuinely non-Markovian robotic manipulation. Unlike conventional long-horizon environments where the workspace state remains persistently visible, \oursbench~enforces strict visual occlusions and temporal delays. In these tasks, critical visual evidence, such as the hidden color of a block, the identity of an object under a cover, or a randomly generated spatial sequence, manifests only transiently during intermediate interactions before becoming completely unobservable. 

To systematically quantify memory capacity, \oursbench~spans 8 complex bimanual manipulation tasks with exceptionally long execution horizons, ranging from an average of 430 to 1,544 steps per episode. The difficulty of each task is explicitly parameterized by $n \in [1, 5]$, which defines the exact number of intermediate keyframe events the robot must autonomously capture and retain to successfully complete the instruction. 

For instance, memory-intensive tasks like \textit{Cover Blocks Hard} ($n=4$) and \textit{Pick Objects in Order} ($n=3$) require the robot to lift opaque covers to inspect hidden attributes, remember them after the covers are closed, and execute subsequent pick-and-place actions based on that stored memory. Similarly, \textit{Press Button Keyframe} requires the robot to read randomized number cards and translate them into a sequential counting and pressing logic. As visualized in Fig.~\ref{fig:memdojo}, these transient, interaction-driven keyframes (highlighted with blue borders) serve as the critical informational bridge between past observations and future actions. 

The comprehensive task statistics, including the average number of steps, the required intermediate keyframe count $n$, and the specific language instructions for all 8 evaluation tasks, are detailed in Table~\ref{memdojo}.

\subsection{Real-world Tasks Details}

\begin{figure*}[t] 
\centering 
\includegraphics[width=0.98\linewidth]{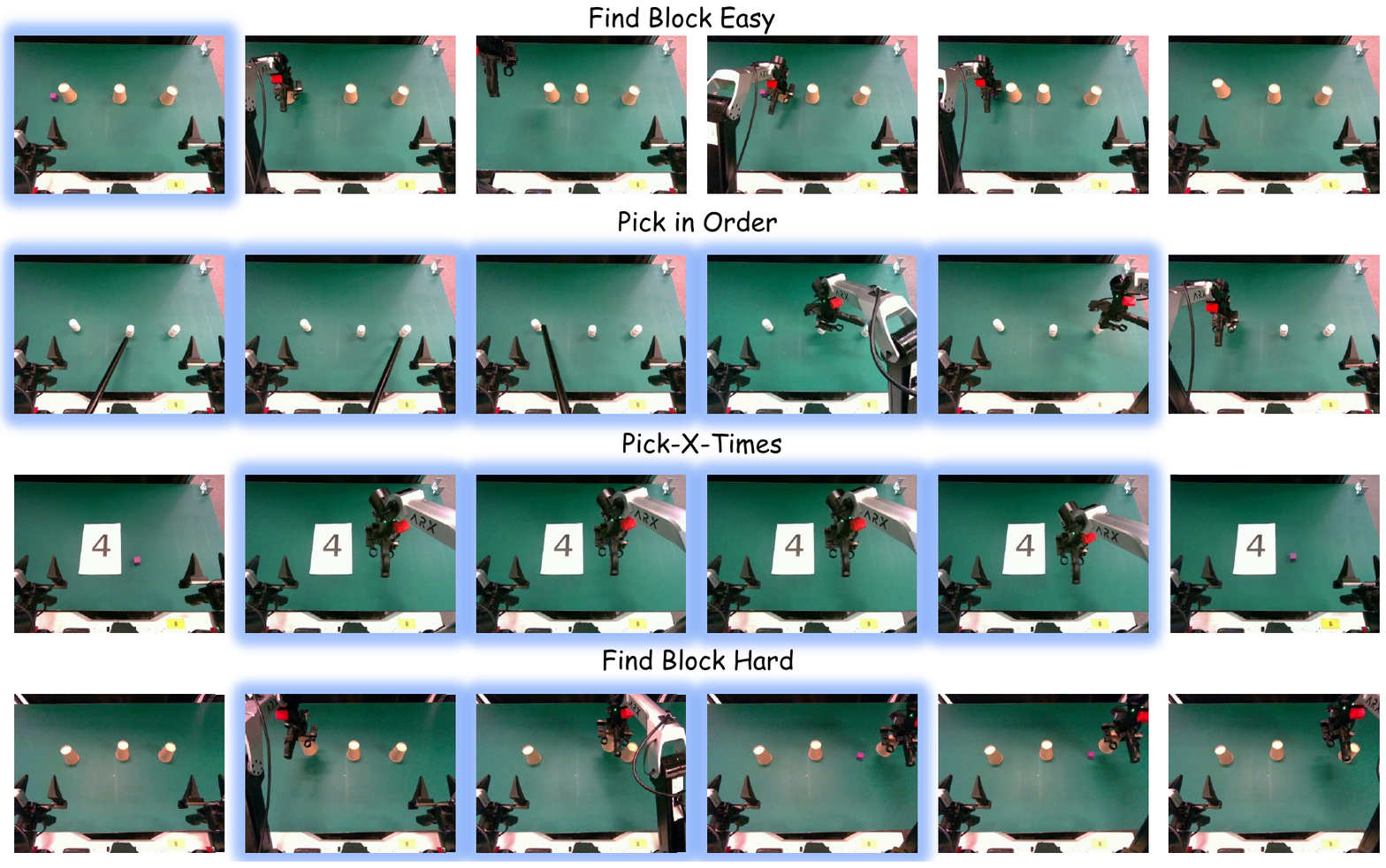} 
\caption{Expanded real-world execution sequences of EventVLA across the four manipulation tasks. The specific task-critical intermediate keyframes, which the policy autonomously captures and commits to memory, are highlighted with blue borders.} 
\label{fig:realworld_tasks} 
\end{figure*}

To supplement the single-frame task overviews provided in Fig.~\ref{fig:real_robot}, Fig.~\ref{fig:realworld_tasks} presents the expanded, step-by-step temporal sequences for the four real-world manipulation tasks. These full execution rollouts illustrate exactly when transient visual evidence emerges during physical interaction. The task-critical intermediate keyframes that the policy must autonomously capture and commit to its dynamic event buffer, such as briefly exposing a hidden block, reading a randomized number, or observing a specific sequence pointed out by a stick, are explicitly highlighted with blue borders. By successfully isolating and retaining these sparse states, EventVLA effectively bridges the temporal gap required for non-Markovian control.

\subsection{Network Architecture and Hyper-parameters}
\label{appen:hyperparas}

\paragraph{Baseline implementation.}
All compared methods are trained using the same task demonstrations and evaluated under the same protocol. Equal-budget sparse-memory baselines use the same maximum memory capacity as \ours~and differ only in their frame-selection strategy. For baselines requiring keyframe supervision, we provide the same Qwen3-VL-generated keyframe annotations used by \ours. The subtask labels required by dual-system baselines are manually annotated following their original formulations.
For the equal-budget frame-selection baselines, Random uniformly samples memory-write steps, Periodic stores one frame every $P=100$ steps, and Visual-change stores a frame when the visual-difference score exceeds a threshold $\delta_{\mathrm{vis}}=10$. The Reactive selector predicts whether the currently observed frame should be stored, without forecasting future keyframes. Oracle directly uses the ground-truth keyframe timestamps. All methods use the same maximum event-buffer capacity $N_{\max}=5$.

For RMBench and \oursbench, our \ours~framework is built upon the open-source QwenOFT~\cite{starvla} architecture, which utilizes Qwen3-VL-4B-Instruct as the foundational Vision-Language Model (VLM). The visual observations are resized to $224 \times 224$ before being processed by the vision encoder.

During the training phase, the entire framework is optimized end-to-end using the AdamW optimizer for 80,000 training steps. We apply a differential learning rate strategy to ensure stable convergence: the pre-trained VLM backbone is fine-tuned with a lower learning rate of $1 \times 10^{-5}$, while the newly initialized components (the action head and the Keyframe Evidence Memory prediction head) are trained with a higher learning rate of $1 \times 10^{-4}$. The action prediction horizon ($H$) is set to 50 steps for all tasks.

\input{table/rmbench_hyper}

To explicitly reflect the different memory demands of our evaluated benchmarks, we configure the memory modules differently. For \textbf{RMBench}, which primarily evaluates foundational visual anchoring without the need for intermediate transient memory, the policy is trained exclusively with initial and short-term visual anchors. The detailed network architecture and training hyper-parameters for RMBench are summarized in Table~\ref{tab:rmbench_hyper}. 

Conversely, for the strictly non-Markovian \textbf{\oursbench}, the full Keyframe Evidence Memory (KEM) module is activated. To ensure early training stability and bridge the train-test distribution shift, we apply a scheduled teacher-to-student curriculum, where the teacher-forcing probability $\alpha$ decays linearly from 1.0 to 0.0 over the training duration. As detailed in Table~\ref{tab:memdojo_hyper}, we also introduce specific hyper-parameters to govern the online memory extraction pipeline. The chunk-wise keyframe predictions are filtered using a commit confidence threshold of $\tau_{\text{commit}} = 0.55$. To enforce rigorous memory sparsity, we apply a 1D Non-Maximum Suppression (NMS) sliding window with a radius of $w = 8$, followed by a temporal cooldown period of $C = 10$ steps between consecutive memory writes. Finally, the dynamic event buffer is bounded by a maximum capacity of $N_{\text{max}} = 5$, managed by a FIFO eviction policy to satisfy real-time computational constraints.

\input{table/memdojo_hyper}

\input{table/real_robot_hyper}

For physical deployment on the real-world robot platform, we adapt our framework to utilize \textbf{$\pi_{0.5}$}~\cite{pi05} as the foundational Vision-Language-Action Model. The action head is configured to predict a 32-dimensional continuous action over a horizon of $H = 50$ steps. During fine-tuning on real-world demonstrations, the entire framework is jointly optimized for 60,000 steps using the AdamW optimizer with a global batch size of 32 in bfloat16 precision. We apply a uniform peak learning rate of $5 \times 10^{-5}$ for both the base VLM and the newly initialized heads, following a cosine decay schedule with 2,000 warm-up steps. To manage the Keyframe Evidence Memory (KEM) module during physical execution, we maintain a commit confidence threshold of $\tau_{\text{commit}} = 0.55$, a maximum event buffer capacity of $N_{\text{max}} = 5$, an NMS temporal window radius of $w = 8$, and a commit cooldown period of $C = 10$. The keyframe loss weight $\lambda$ is set to 0.1, alongside a scheduled teacher-to-student curriculum where $\alpha$ decays linearly from 1.0 to 0.0. The comprehensive network architecture and training details for the real-world tasks are summarized in Table~\ref{tab:real_robot_hyper}.

\section{Extended Experimental Results and Analysis}

\input{table/rmbench}
\subsection{Detailed Per-Task Breakdown on RMBench}
Due to space limits in the main text, we present the comprehensive task-level breakdown for all baseline and ablation models on the RMBench suite in Table~\ref{benchmark:rmbench}, which is the task-specific expanded version of Table~\ref{benchmark:rmbench_abbr}. 

The detailed breakdown clearly illustrates that our streamlined configuration, EventVLA (visual anchors only), significantly outperforms both non-memory and prior memory-augmented baselines across the vast majority of tasks. Notably, on memory-intensive structural manipulation tasks such as \textit{Rearrange Blocks} (96\%), \textit{Put Back Block} (95\%), \textit{Swap Blocks} (96\%), and \textit{Cover Blocks} (97\%), EventVLA achieves near-perfect success rates. This demonstrates that our rule-based visual anchoring mechanism effectively captures the persistent spatial layouts and fixed motion styles required for conventional memory-oriented scenarios without the need for complex state compression. Furthermore, the ablation variants (w/o initial and w/o short-term) show severe performance degradation across almost all tasks, confirming that both the initial global spatial reference and short-term motion cues are indispensable components of the visual anchors.

\subsection{Extended Ablation Analysis and Inference Efficiency}
\label{appendix:abla}
\input{table/rebuttal_foresight}

\textbf{Extended Ablation Analysis.}
To deeply understand the contributions of individual design choices within the Keyframe Evidence Memory (KEM) module, we expand upon the ablation studies conducted on the \oursbench~suite (summarized in the main text Sec.~\ref{sec:eventvla_abla} and Table~\ref{benchmark:memdojo}).
\begin{itemize}
    \item \textbf{Memory Representation (Explicit Images vs. Implicit Bank):} In the implicit memory bank variant, captured keyframes are aggregated into a compressed latent embedding rather than appended as explicit raw images. When handling complex tasks that demand the retention of multiple distinct events (e.g., $n \ge 3$), squeezing disparate historical features into a single latent vector creates a severe information bottleneck. Explicit raw image concatenation avoids this lossy compression, providing complete, lossless contextual evidence for the VLA's multi-frame attention mechanism.
    \item \textbf{Supervision Strategy (Soft Labels vs. Hard Labels):} Physical keyframe events naturally span continuous temporal windows. Replacing our raised cosine soft labels with strict binary targets induces extreme label sparsity and heavily penalizes valid adjacent frames. This rigid supervision destabilizes the predictive head, ultimately causing it to fail in triggering essential memory writes. Soft labels provide the necessary temporal tolerance for robust event capture in environments with execution variance.
    \item \textbf{Buffer Management (The Necessity of NMS and Capacity):} Without the 1D Non-Maximum Suppression (NMS) post-processing algorithm, redundant adjacent frames rapidly flood the bounded dynamic event buffer. Conversely, a strictly minimal buffer (e.g., $N_{\text{max}}=2$) inherently lacks the structural capacity required for complex, multi-stage tasks. Both scenarios lead to premature buffer saturation and trigger early FIFO eviction, which mistakenly discards foundational historical evidence (such as the first observed hidden color) before it can be utilized. This underscores that NMS-driven event sparsity and adequate memory capacity are both vital.
    \item \textbf{Foresight Horizon:}
    To isolate keyframe foresight from action-chunk length, we fix the action horizon at $H=50$ and vary only the keyframe foresight horizon $H_k$.
    As presented in Table~\ref{tab:rebuttal_foresight},
    the reactive selector ($H_k=0$) achieves 28.9\% on \oursbench, while increasing $H_k$ from 1 to 15, 30, and 50 progressively improves performance from 28.2\% to 36.8\%, 45.4\%, and 75.2\%, respectively. This controlled comparison confirms the benefit of longer keyframe foresight independently of action-chunk length.
    When varying $H_k$, we keep the action tokens and action loss unchanged and mask keyframe predictions beyond the first $H_k$ future steps.
    
\end{itemize}

\input{table/inference_cost}

\textbf{Inference Efficiency.}
To verify that \ours~can be effectively deployed on physical robots, we evaluate its inference cost. Table~\ref{tab:inference_cost} reports the average per-chunk latency and GPU memory usage measured on an NVIDIA A100 GPU with batch size 1.

As shown in Table~\ref{tab:inference_cost}, \ours~requires 0.42 seconds per action chunk and approximately 7 GB of GPU memory on an NVIDIA A100 GPU. In comparison, the dual-system MemER and $\pi_{MEM}$ baselines require 1.43/1.08 seconds per chunk and approximately 24 GB of GPU memory, respectively. These results support the inference-efficiency advantage of the end-to-end design.

\section{Qualitative Visualizations}

\subsection{Simulation Rollouts in \oursbench}
\begin{figure*}[t] \centering \includegraphics[width=0.98\linewidth]{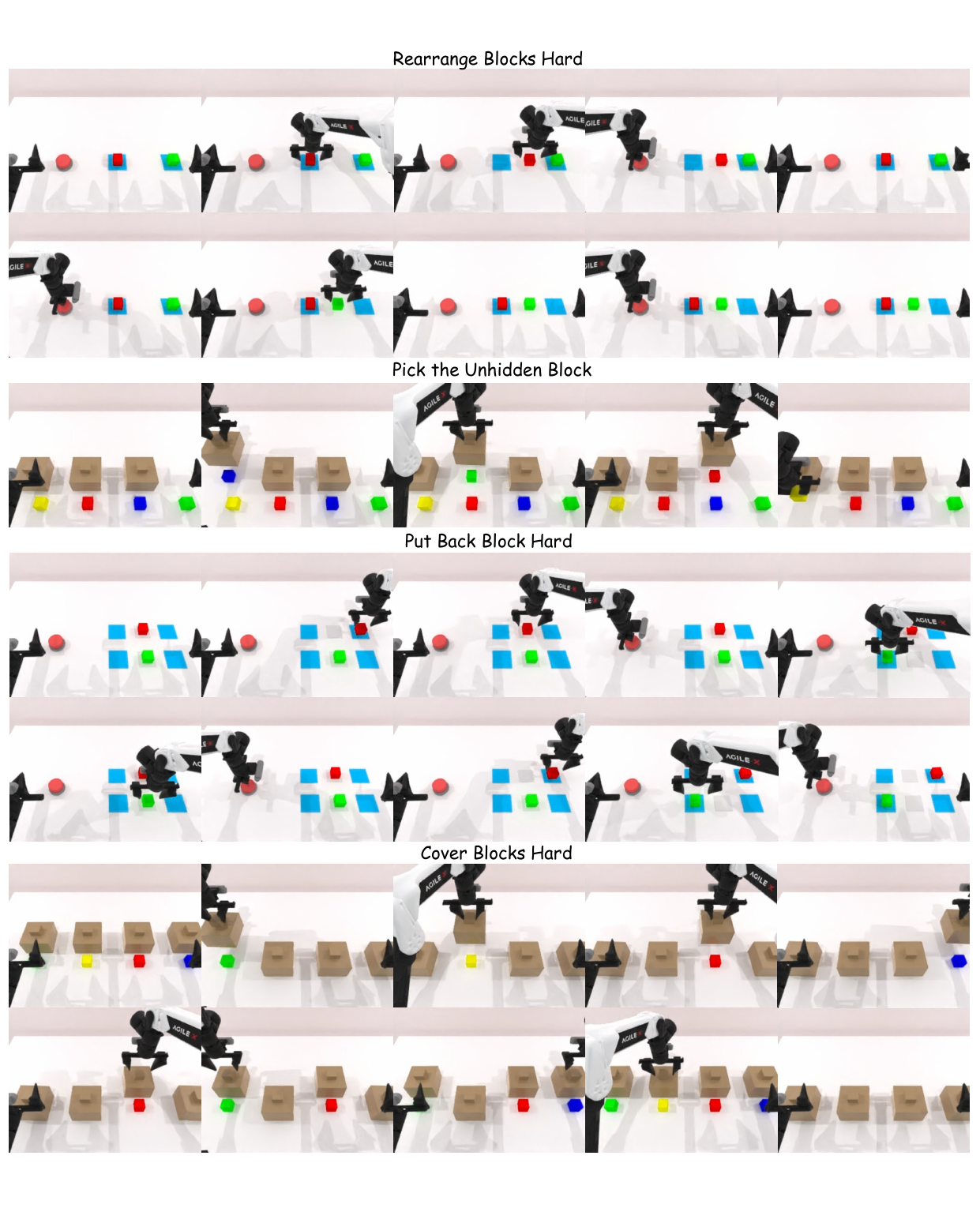} \caption{ Qualitative rollouts of EventVLA on four \oursbench~simulation tasks: \textit{Rearrange Blocks Hard}, \textit{Pick the Unhidden Block}, \textit{Put Back Block Hard}, and \textit{Cover Blocks Hard}. } \label{fig:memdojo_rollout_part1} \end{figure*}

\begin{figure*}[t] \centering \includegraphics[width=0.98\linewidth]{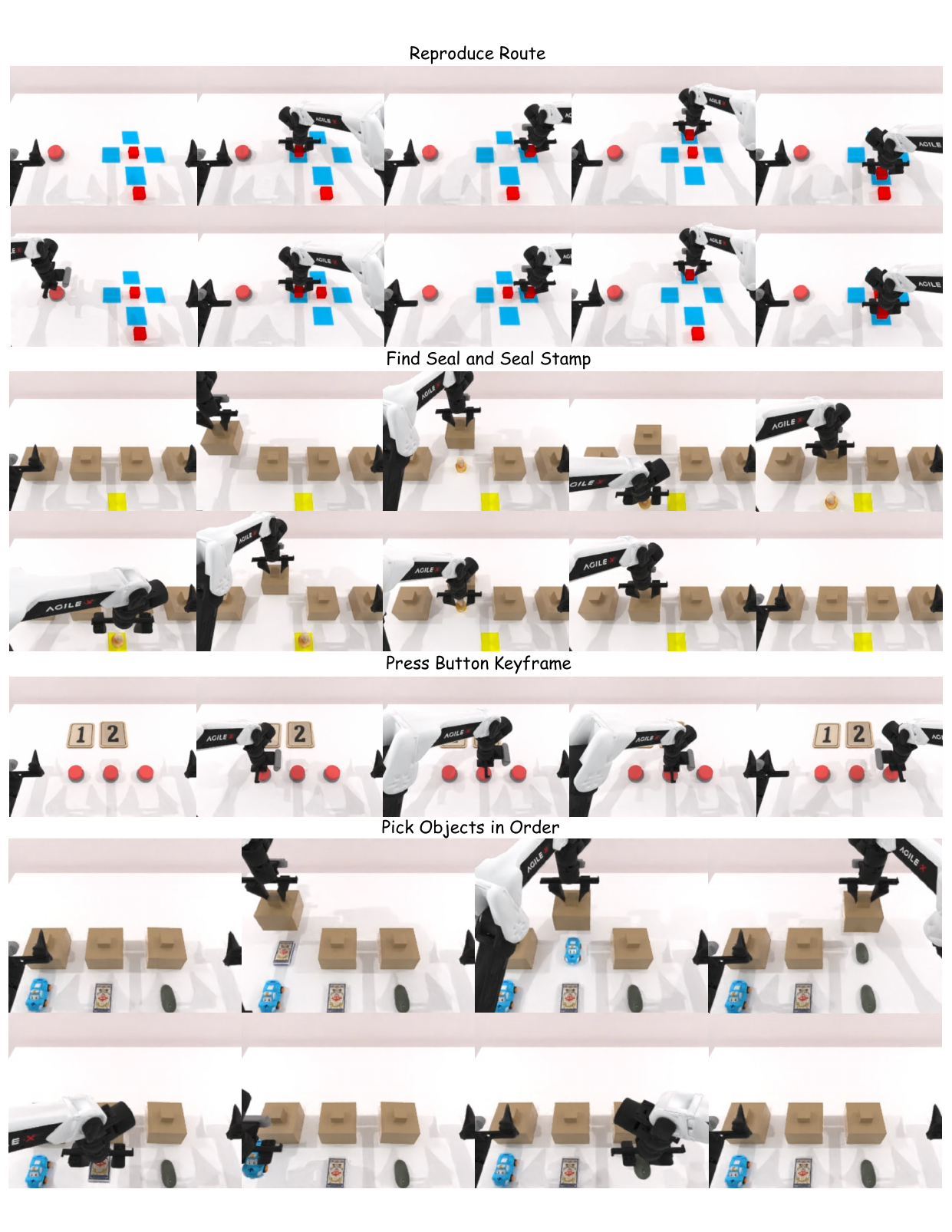} \caption{ Qualitative rollouts of EventVLA on the remaining four \oursbench~simulation tasks: \textit{Press Button Keyframe}, \textit{Pick Objects in Order}, \textit{Find Seal and Seal Stamp}, and \textit{Reproduce Route}. } \label{fig:memdojo_rollout_part2} \end{figure*}

To provide an intuitive understanding of EventVLA's dynamic memory scheduling and execution process, we visualize the qualitative rollouts across all 8 strictly non-Markovian tasks in the \oursbench~benchmark. Figure~\ref{fig:memdojo_rollout_part1} illustrates the successful execution sequences for four memory-intensive tasks: \textit{Rearrange Blocks Hard}, \textit{Pick the Unhidden Block}, \textit{Put Back Block Hard}, and \textit{Cover Blocks Hard}. Figure~\ref{fig:memdojo_rollout_part2} demonstrates the execution pipelines for the remaining four tasks: \textit{Press Button Keyframe}, \textit{Pick Objects in Order}, \textit{Find Seal and Seal Stamp}, and \textit{Reproduce Route}.

Across these diverse scenarios, the visualizations clearly highlight how the KEM module proactively triggers sparse memory writes the exact moment transient visual evidence emerges (e.g., observing the hidden color of a block immediately after lifting an opaque cover, or reading a randomized number). By locking these critical intermediate states into the event buffer before they become unobservable, EventVLA effectively bridges the temporal gap and seamlessly guides the subsequent long-horizon manipulation.

\subsection{Real-World Robot Execution Sequences}

\begin{figure*}[t] \centering \includegraphics[width=0.98\linewidth]{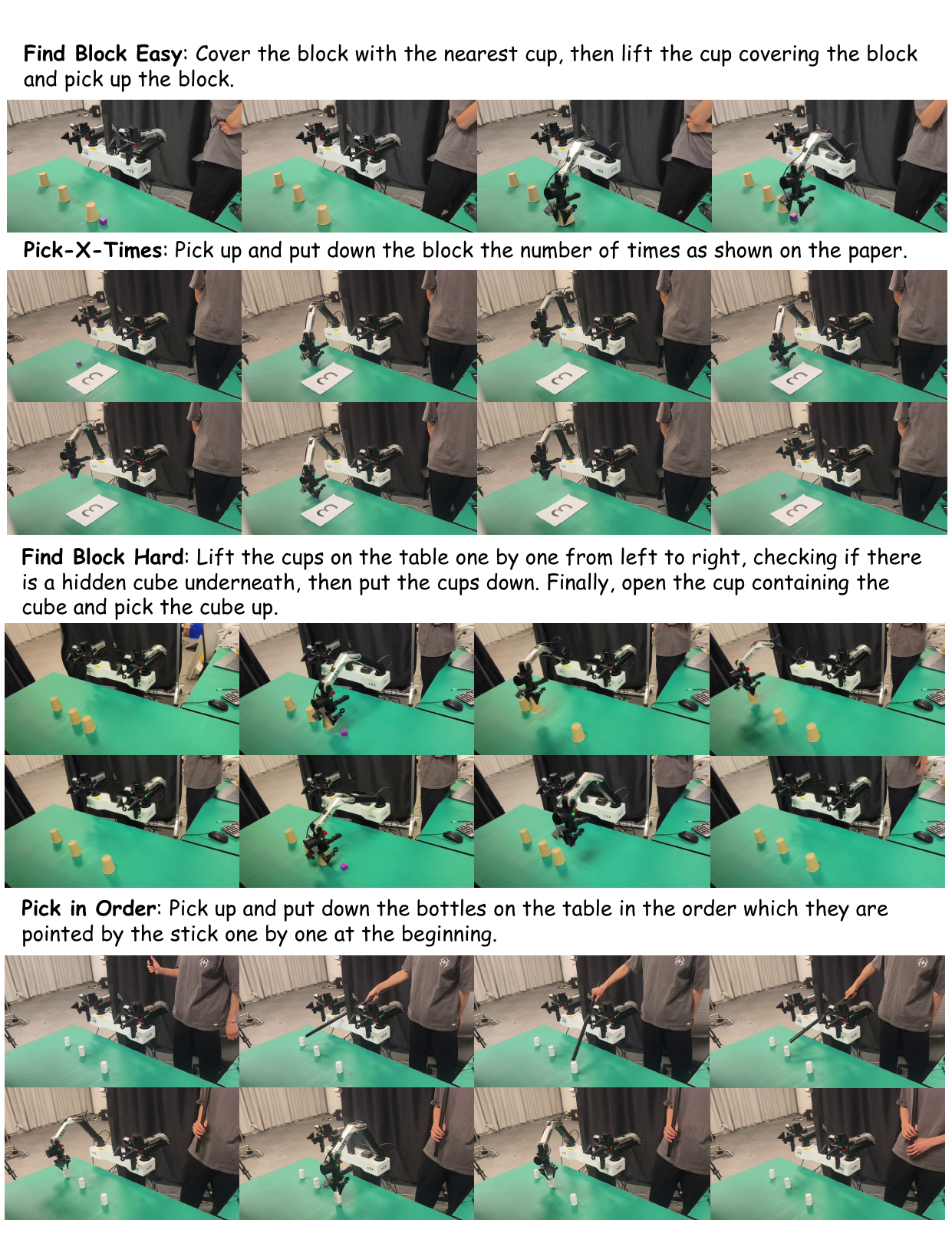} \caption{ Qualitative real-world robot execution sequences of EventVLA on four tasks: \textit{Find Block Easy}, \textit{Find Block Hard}, \textit{Pick-X-Times}, and \textit{Pick in Order}. } \label{fig:realworld_rollout} \end{figure*}

To further validate the practical efficacy of our framework, we provide qualitative execution sequences of EventVLA deployed on the real-world ARX ACONE bimanual robot. Figure~\ref{fig:realworld_rollout} showcases the successful completion of four memory-intensive manipulation tasks: \textit{Find Block Easy}, \textit{Pick-X-Times}, \textit{Find Block Hard}, and \textit{Pick in Order}. 

The visualizations demonstrate that EventVLA can robustly capture and retain critical intermediate visual cues despite real-world occlusions and randomized spatial placements. Whether reading a randomized number from a paper to dictate counting logic, or observing a stick pointing at bottles to memorize an in-context sequence, the policy successfully utilizes its sparse visual evidence memory to execute complex, multi-stage physical tasks, exhibiting both strong non-Markovian remembering and spatial generalization capabilities.

\end{document}

%% file: table/rmbench_abbr.tex
\begin{wraptable}{r}{0.35\textwidth}
\centering
\renewcommand{\arraystretch}{0.95}
\setlength{\tabcolsep}{3pt}
\caption{Overall average success rates on RMBench. Detailed per-task breakdowns are in Appendix Table~\ref{benchmark:rmbench}.}
\tiny
\begin{tabular}{lc}
\toprule
\textbf{Methods} & \textbf{Average (\%)} \\
\midrule
\multicolumn{2}{l}{\textit{Non Memory-based VLAs}} \\
DP & 5.8 \\
ACT & 5.9 \\
$\pi_{0.5}$ & 10.4 \\
X-VLA & 9.8 \\
QwenOFT & 5.6 \\
\midrule
\multicolumn{2}{l}{\textit{Dual-system Memory-VLAs}} \\
MemER & 8.7 \\
Mem-0 & 42.0 \\
\midrule
\multicolumn{2}{l}{\textit{End-to-end Memory-VLAs}} \\
MemoryVLA (OpenVLA) & 19.4 \\
MemoryVLA (QwenOFT) & 41.7 \\
\midrule
\multicolumn{2}{l}{\textit{\ours \& Ablations}} \\
EventVLA (w/o initial) & 33.7 \\
EventVLA (w/o short-term) & 23.8 \\
\textbf{EventVLA (VA only)} & \textbf{67.8} \\
\bottomrule
\end{tabular}
\vspace{-2mm}
\label{benchmark:rmbench_abbr}
\end{wraptable}

%% file: table/rmbench_new.tex
\begin{table}[t]
\centering
\renewcommand{\arraystretch}{0.95}
\setlength{\tabcolsep}{3pt}
\scriptsize
\caption{{\oursbench~benchmark results.} (\colorbox{blue!15}{{\textbf{Bold}}}: best; {\underline{Underlined}}: second-best).
}
\resizebox{\columnwidth}{!}{%
\begin{tabular}{l|ccccccccc}
\toprule

\multirow{3}{*}{Tasks}  & \multicolumn{1}{c}{n=1} &
        \multicolumn{1}{c}{n=2} &
        \multicolumn{2}{c}{n=3}
        & \multicolumn{3}{c}{n=4} 
        & \multicolumn{1}{c}{n=5} 
        & \multirow{3}{*}{\textit{\makecell{Total \\ average}}}\\

\cmidrule(lr){2-2} \cmidrule(lr){3-3} \cmidrule(lr){4-5} \cmidrule(lr){6-8} \cmidrule(lr){9-9}

 & \makecell{Rearrange\\ Blocks Hard} & \makecell{Put Back \\Block Hard} & \makecell{Pick Objects \\in Order} & \makecell{Pick the \\Unhidden Block}  & \makecell{Cover\\Blocks Hard} & \makecell{Find Seal\\ Stamp}  & \makecell{Reproduce\\ Route} &  \makecell{Press Button\\ Keyframe} &  \\
\midrule
\rowcolor{gray!15} \multicolumn{10}{l}{$\blacktriangledown$ \emph{Non Memory-based Vision-language-action Models:}}   \\
$\pi_{0.5}$ & 20\% &	19\% &	1\% &	14\% &	0\% &	8\% &	0\% &	0\% &	7.8\% \\
QwenOFT &3\%&	\underline{26\%}&	0\%&	0\%&	0\%&	0\%&	0\%&	1\%	&3.8\%\\
\rowcolor{gray!15} \multicolumn{10}{l}{$\blacktriangledown$ \emph{Dual-system Memory-based Vision-language-action Models:}}   \\
MemER &32\%	&4\%	&\underline{12\%}	&2\%	&0\%	&\underline{26\%}	&\underline{3\%}	&5\%	&10.5\%
\\
Mem-0 & 0\% & 0\% & 0\% & 0\% & 0\% & 0\% & 0\% & 0\% & 0.0\% \\
\rowcolor{gray!15} \multicolumn{10}{l}{$\blacktriangledown$ \emph{End-to-end Memory-based Vision-language-action Models:}}   \\
MemoryVLA (OpenVLA) &12\%	&0\%&	0\%&	1\%	&0\%	&10\%	&2\%	&14\%	&4.9\%\\
MemoryVLA (QwenOFT) &\underline{39\%}	&0\%	&1\%	&9\%	&\underline{1\%}	&11\%	&0\%	&\underline{25\%}	&10.8\%\\
EventVLA (VA only) &\colorbox{blue!15}{\textbf{62\%}}	&13\%	&5\%	&\underline{20\%}	&0\%	&\underline{26\%}	&0\%	&18\%	&\underline{18.0\%}\\
EventVLA (VA+KEM) &\colorbox{blue!15}{\textbf{62\%}}	&\colorbox{blue!15}{\textbf{93\%}}	&\colorbox{blue!15}{\textbf{90\%}}	&\colorbox{blue!15}{\textbf{54\%}}	&\colorbox{blue!15}{\textbf{94\%}}	&\colorbox{blue!15}{\textbf{63\%}}	&\colorbox{blue!15}{\textbf{98\%}}	&\colorbox{blue!15}{\textbf{48\%}}	&\colorbox{blue!15}{\textbf{75.2\%}}  \\
\midrule
\color{gray} EventVLA (implicit memory bank) & \color{gray} 51\%	&\color{gray} 9\%	&\color{gray} 16\%	&\color{gray} 37\%	&\color{gray} 1\%	&\color{gray} 68\%	&\color{gray} 2\%	&\color{gray} 15\%	&\color{gray} 24.9\%\\
\color{gray} EventVLA (hard label) & \color{gray} 59\%	& \color{gray} 77\%	& \color{gray} 28\%	& \color{gray} 62\%	& \color{gray} 85\%	& \color{gray} 36\%	& \color{gray}6\% &	\color{gray}37\% & \color{gray}48.8\%	\\
\color{gray} EventVLA (w/o NMS) & \color{gray} 62\%	&\color{gray} 93\%	&\color{gray} 49\%	&\color{gray} 36\%	&\color{gray} 10\%	&\color{gray} 35\%	&\color{gray} 97\%	&\color{gray} 45\%	&\color{gray} 53.4\% \\

\color{gray} EventVLA ($N_{\text{max}}=2$) & \color{gray}51\%	& \color{gray}35\%	& \color{gray}28\%	& \color{gray}33\%	& \color{gray}39\%	&\color{gray}53\%	&\color{gray}0\%	&\color{gray}17\%	&\color{gray}32.0\% \\

\bottomrule

\end{tabular}%
}
\vspace{-4mm}
\label{benchmark:memdojo}
\end{table}

%% file: table/robotwin50.tex
\begin{wraptable}{r}{0.45\textwidth}
\centering
\renewcommand{\arraystretch}{0.95}
\setlength{\tabcolsep}{3pt}
\scriptsize
\caption{\textbf{RoboTwin2.0 benchmark results.}
\ours~outperforms its baseline foundation model QwenOFT on Markovian tasks.
}
\resizebox{\linewidth}{!}{%
\begin{tabular}{l|cccccc}
\toprule
Tasks & $\pi_0$ & $\pi_{0.5}$ & X-VLA & \makecell{QwenFast} & \makecell{QwenOFT} & EventVLA \\
\midrule

Easy & 65.9\% & 82.7\% & 72.8\% & 72.5\% & 80.0\% & 83.8\%\\

Hard & 58.4\% & 76.8\% &72.8\% & 83.2\% & 78.0\%  & 81.6\%\\

\bottomrule

\end{tabular}%
}
\vspace{-2mm}
\label{benchmark:robotwin2.0}
\end{wraptable}

%% file: table/rebuttal_memory.tex
\begin{table*}[t]
\centering
\caption{
Controlled comparison of memory construction strategies and model components.
All frame-selection baselines use the same backbone and memory budget.
Real-world results for the controlled ablations are averaged across multiple task configurations and random seeds.
}
\label{tab:rebuttal_memory}
\resizebox{0.92\textwidth}{!}{
\begin{tabular}{lccccc}
\toprule
Method / Memory Strategy
& Foresight
& RMBench
& RoboTwin-MeM
& RoboMME
& Real \\
\midrule
Random            & $\times$ & --   & 10.1 & --   & 1.3  \\
Periodic          & $\times$ & --   & 34.2 & --   & 37.1 \\
Visual-change     & $\times$ & --   & 44.6 & --   & 22.6 \\
Reactive selector & $\times$ & --   & 28.9 & --   & 30.4 \\
\midrule
VA only           & $\times$     & 67.8 & 18.0 & 16.4 & 25.0 \\
KEM only          & \checkmark   & 30.3 & 50.7 & 28.3 & 53.6 \\
VA+KEM            & \checkmark   & 66.7 & 75.2 & 32.1 & 78.8 \\
Oracle keyframes  & Oracle       & 67.4 & 76.3 & 40.5 & 81.8 \\
\bottomrule
\end{tabular}
}
\end{table*}

%% file: table/memdojo_task.tex
\begin{table*}[ht]
\centering
\renewcommand{\arraystretch}{1.25}
\setlength{\tabcolsep}{5pt}
\small 
\caption{{Detailed statistics of our proposed \oursbench~Benchmark.}}
\begin{tabularx}{\textwidth}{l|ccc|X} 
\toprule
\textbf{Task Name}
& \textbf{Episodes}
& \textbf{\makecell{Avg.\\\#Steps}}
& \textbf{\makecell{Intermediate\\Keyframes}}
& \textbf{Task Instruction} \\

\midrule

Press Button Keyframe
& 50 & 430 & [2,5]
& Read the two number cards, press the left button as many times as the left card shows, press the middle button as many times as the right card shows, then press the right button once. \\

\rowcolor{gray!15}
Pick the Unhidden Block
& 50 & 699 & 3
& Open the covers one by one to identify the hidden colors, close them after inspection, then pick up the visible block whose color is not hidden. \\

Rearrange Blocks Hard
& 50 & 879 & 1
& Move a chosen block from its mat to the center and press the button, return the same block to its mat and press again, then move the other block to the center and press once more. \\

\rowcolor{gray!15}
Pick Objects in Order
& 50 & 1124 & 3
& Open the covers one by one to observe the objects inside, close them after inspection, then pick up the objects in the observed order. \\

Find Seal and Seal Stamp
& 50 & 1338 & [1,4]
& Open the covers one by one to find and take out the seal, close the cover after inspection, stamp with it, then return it under its original cover. \\

\rowcolor{gray!15}
Reproduce Route
& 50 & 1417 & 4
& Move the center red block to the four blue pads in a random order, returning it to the center. Then use the outside red block to repeat the same pad order. \\

Put Back Block Hard
& 50 & 1468 & 2
& For each row, move the center block to a randomly selected outer pad in the same row, move the arm back, return the block to the center, move the arm back again, and press the button. Finally, move both blocks back to the same outer pads they first visited, then press the button. \\

\rowcolor{gray!15}
Cover Blocks Hard
& 50 & 1544 & 4
& Open the covers one by one, close them after inspection, then reopen them in the order: red, green, blue, yellow. \\

\bottomrule
\end{tabularx} 
\vspace{-2mm}
\label{memdojo}
\end{table*}

%% file: table/rmbench_hyper.tex
\begin{wraptable}{r}{0.55\textwidth}
\centering
\renewcommand{\arraystretch}{0.95}
\setlength{\tabcolsep}{3pt}
\scriptsize
\caption{Network Architecture and Training Hyper-parameters for RMBench.}
\resizebox{\linewidth}{!}{%
\begin{tabular}{lc}
\toprule
\textbf{Configurations} & \textbf{Values} \\
\midrule
\multicolumn{2}{l}{\textit{Network Architecture}} \\
Base VLM & Qwen3-VL-4B-Instruct \\
Action Model Type & Optimized Fine-Tuning (OFT) \\
Action Dimension & 14 \\
Action Horizon ($H$) & 50 \\
Image Resolution & $224 \times 224$ \\
\midrule
\multicolumn{2}{l}{\textit{Training Hyper-parameters}} \\
Optimizer & AdamW \\
Training Steps & 80,000 \\
Base VLM Learning Rate & $1 \times 10^{-5}$ \\
Action Head Learning Rate & $1 \times 10^{-4}$ \\
Per-Device Batch Size & 4 \\
Gradient Accumulation Steps & 1 \\
Memory Module Status & Visual Anchors Only \\
Visual Anchors $A_t$ & $o_0, o_{t-30}, o_{t-15}$ \\
\bottomrule
\end{tabular}
}
\vspace{-2mm}
\label{tab:rmbench_hyper}
\end{wraptable}

%% file: table/memdojo_hyper.tex
\begin{table}[htbp]
\centering
\small
\caption{Network Architecture and KEM Hyper-parameters for \oursbench.}
\label{tab:memdojo_hyper}
\vspace{2mm}
\begin{tabular}{lc}
\toprule
\textbf{Configurations} & \textbf{Values} \\
\midrule
\multicolumn{2}{l}{\textit{Network Architecture \& Basic Training}} \\
Base VLM & Qwen3-VL-4B-Instruct \\
Action Model Type & Optimized Fine-Tuning (OFT) \\
Action Horizon ($H$) & 50 \\
Optimizer & AdamW \\
Training Steps & 80,000 \\
Base VLM Learning Rate & $1 \times 10^{-5}$ \\
KEM \& Action Head Learning Rate & $1 \times 10^{-4}$ \\
Per-Device Batch Size & 4 \\
Gradient Accumulation Steps & 1 \\
\midrule
\multicolumn{2}{l}{\textit{Keyframe Evidence Memory (KEM) Settings}} \\
Memory Module Status & Full EventVLA (VA + KEM) \\
Visual Anchors $A_t$ & $o_0, o_{t-30}, o_{t-15}$ \\
Teacher-Forcing Annealing ($\alpha$) & Linear decay (1.0 $\rightarrow$ 0.0) \\
Commit Confidence Threshold ($\tau_{\text{commit}}$) & 0.55 \\
Max Event Buffer Size ($N_{\text{max}}$) & 5 \\
NMS Temporal Window Radius ($w$) & 8 \\
Commit Cooldown Period ($C$) & 10 \\
Keyframe Loss Weight ($\lambda$) & 0.1 \\
\bottomrule
\end{tabular}
\end{table}

%% file: table/real_robot_hyper.tex
\begin{table}[htbp]
\centering
\small
\caption{Network Architecture and KEM Hyper-parameters for real-robot.}
\label{tab:real_robot_hyper}
\vspace{2mm}
\begin{tabular}{lc}
\toprule
\textbf{Configurations} & \textbf{Values} \\
\midrule
\multicolumn{2}{l}{\textit{Network Architecture \& Basic Training}} \\
Base VLM & PaliGemma ($\pi_{0.5}$) \\
Image Resolution & $224 \times 224$ \\
Text Sequence Length & 200 \\
Action Horizon ($H$) & 50 \\
Action Dimension & 32 \\
Optimizer & AdamW \\
Optimizer Hyper-parameters & $\beta_1=0.9$, $\beta_2=0.95$, $eps=1e-8$ \\
Weight Decay & 0.01 \\
Training Steps & 60,000 \\
Warm-up Steps & 2,000 \\
Base VLM Learning Rate & $5 \times 10^{-5}$ \\
KEM \& Action Head Learning Rate & $5 \times 10^{-5}$ \\
Minimum Learning Rate & $5 \times 10^{-6}$ \\
Learning Rate Schedule & cosine decay with minimum LR \\
Global Batch Size & 32 \\
Numerical Precision & bfloat16 \\
\midrule
\multicolumn{2}{l}{\textit{Keyframe Evidence Memory (KEM) Settings}} \\
Memory Module Status & Full EventVLA (VA + KEM) \\
Visual Anchors $A_t$ & $o_0, o_{t-60}, o_{t-40}, o_{t-20}$ \\
Teacher-Forcing Annealing ($\alpha$) & linear decay (1.0 $\rightarrow$ 0.0) \\
Commit Confidence Threshold ($\tau_{\text{commit}}$) & 0.55 \\
Max Event Buffer Size ($N_{\text{max}}$) & 5 \\
NMS Temporal Window Radius ($w$) & 8 \\
Commit Cooldown Period ($C$) & 10 \\
Keyframe Loss Weight ($\lambda$) & 0.1 \\
\bottomrule
\end{tabular}
\end{table}

%% file: table/rmbench.tex
\begin{table}[ht]
\centering
\renewcommand{\arraystretch}{0.95}
\setlength{\tabcolsep}{3pt}
\scriptsize
\caption{{RMBench benchmark results.} (\colorbox{blue!15}{{\textbf{Bold}}}: best; {\underline{Underlined}}: second-best).
}
\resizebox{\columnwidth}{!}{%
\begin{tabular}{l|cccccccccc}
\toprule
Tasks & \makecell{Observe and \\Pick Up} & \makecell{Rearrange\\ Blocks} & \makecell{Put Back \\Block} & \makecell{Swap\\ Blocks}  & \makecell{Swap \\T} & \makecell{Battery\\ Try}  & \makecell{Blocks\\ Ranking Try} & \makecell{Cover\\Blocks} & \makecell{Press\\ Button} & \textit{\makecell{Total\\ average}} \\
\midrule
\rowcolor{gray!15} \multicolumn{11}{l}{$\blacktriangledown$ \emph{Non Memory-based Vision-language-action Models:}}   \\
DP & 1\%	&0\%	&0\%	&11\%	&20\%	&10\%	&10\%&	0\%	&0\%	&5.8\%\\
ACT &1\%	&29\%	&0\%	&2\%	&2\%	&19\%	&0\%	&0\%	&0\%	&5.9\%\\
$\pi_{0.5}$ &9\%	&13\%	&11\%	&24\%	&15\%	&16\%	&6\%	&0\%	&0\%	&10.4\%\\
X-VLA &9\%	&13\%	&18\%	&16\%	&3\%	&26\%	&1\%	&2\%	&0\%	&9.8\%\\
QwenOFT &0\% &0\% &0\% &0\% &0\% &14\% & 37\% &0\% &0\% & 5.6\%\\
\rowcolor{gray!15} \multicolumn{11}{l}{$\blacktriangledown$ \emph{Dual-system Memory-based Vision-language-action Models:}}   \\
MemER & 7\%	&17\%	&0\%	&14\%	&7\%	&27\%	&0\%	&6\%	&0\%	&8.7\% \\
Mem-0 &4\%&	\underline{89\%}&	\underline{90\%}&	67\%&	14\%&	28\%&	18\%&	68\%&	0\%&	\underline{42.0}\%\\
\rowcolor{gray!15} \multicolumn{11}{l}{$\blacktriangledown$ \emph{End-to-end Memory-based Vision-language-action Models:}}   \\
MemoryVLA (OpenVLA) & 0\% & 22\% & 50\% & 17\% & 9\% & 25\% & 12\% & 40\% & 0\% & 19.4\%\\
MemoryVLA (QwenOFT) &2\%	&53\%	&81\%	&\underline{76\%}	&9\%	&33\%	&53\%	&\underline{69\%}	&0\%	&41.7\%\\
EventVLA (w/o initial) &10\%	&64\%	&63\%&	16\%	&8\%	&\colorbox{blue!15}{\textbf{39\%}}	&\colorbox{blue!15}{\textbf{87\%}}	&15\%	&\underline{2\%}	&33.7\%\\
EventVLA (w/o short-term) &\underline{15\%}	&34\%&	20\%&	18\%&	\colorbox{blue!15}{\textbf{94\%}}&	16\%&	14\%&	4\%&	0\%	&23.8\%\\
EventVLA (visual anchors only) &\colorbox{blue!15}{\textbf{21\%}}&	\colorbox{blue!15}{\textbf{96\%}}&	\colorbox{blue!15}{\textbf{95\%}}&	\colorbox{blue!15}{\textbf{96\%}}&	\underline{87\%}&	\underline{35\%}&	\underline{81\%}&	\colorbox{blue!15}{\textbf{97\%}}&	\colorbox{blue!15}{\textbf{3\%}}&	 \colorbox{blue!15}{\textbf{67.8\%}}\\

\bottomrule

\end{tabular}%
}
\vspace{-2mm}
\label{benchmark:rmbench}
\end{table}

%% file: table/rebuttal_foresight.tex
\begin{table}[t]
\centering
\caption{
Ablations on keyframe foresight and annotation supervision.
The action horizon is fixed at $H=50$ in all foresight experiments;
only the keyframe foresight horizon $H_k$ is varied.
}
\label{tab:rebuttal_foresight}
\resizebox{0.78\linewidth}{!}{
\begin{tabular}{llcc}
\toprule
Analysis & Setting & RoboTwin-MeM & Real \\
\midrule
\multirow{5}{*}{Foresight}
& Reactive ($H_k=0$) & 28.9 & 30.4 \\
& $H_k=1$            & 28.2 & 31.1 \\
& $H_k=15$           & 36.8 & 42.9 \\
& $H_k=30$           & 45.4 & 50.1 \\
& $H_k=50$           & 75.2 & 78.8 \\
\midrule
\multirow{2}{*}{Annotation}
& Simulator GT / Human & 76.3 & 77.7 \\
& Qwen3-VL              & 75.2 & 78.8 \\
\bottomrule
\end{tabular}
}
\end{table}

%% file: table/inference_cost.tex
\begin{table}[t]
\centering
\caption{
Real-world success rate and inference cost measured on an NVIDIA A100 GPU with batch size 1. Latency includes all model components required to generate one action chunk.
}
\label{tab:inference_cost}
\resizebox{0.72\linewidth}{!}{
\begin{tabular}{lccc}
\toprule
Method & Real Success & Latency & GPU Memory \\
       & (\%) & (s/chunk) & (GB) \\
\midrule
MemER~\cite{memer}       & 32.3 & 1.43 & $\sim$24 \\
$\pi_{MEM}$~\cite{pimem} & 37.5 & 1.08 & $\sim$24 \\
\ours                    & 78.8 & 0.42 & $\sim$7  \\
\bottomrule
\end{tabular}
}
\end{table}